\documentclass[journal]{IEEEtran}

\ifCLASSINFOpdf
\else
\fi

\usepackage{graphicx}
\usepackage{amsmath}
\usepackage{cite}
\usepackage{enumerate}
\usepackage{cases}
\usepackage{multirow}
\usepackage{color}
\usepackage{bm}
\usepackage{booktabs}
\usepackage{hyperref}

\begin{document}

\title{Co-saliency Detection for RGBD Images Based on Multi-constraint Feature Matching and Cross Label Propagation}

\author{Runmin Cong,~\IEEEmembership{Student Member,~IEEE,} Jianjun Lei,~\IEEEmembership{Senior Member,~IEEE,} Huazhu Fu,\par
        Qingming Huang,~\IEEEmembership{Senior Member,~IEEE,} Xiaochun Cao,~\IEEEmembership{Senior Member,~IEEE,}
        and Chunping Hou
\thanks{Manuscript received September 08, 2016; revised May 13, 2017 and September 30, 2017; accepted
October 01, 2017. This work was supported in part by the National Natural Science Foundation of China (No. 61722112, 61520106002, 61731003, 61332016, 61620106009, U1636214, 61602345), and  National Key R\&D Program of China (No. 2017YFB1002900). (\emph{Corresponding author: J. Lei})}
\thanks{R. Cong,  J. Lei, and C. Hou are with the School of Electrical and Information Engineering, Tianjin University, Tianjin 300072, China (e-mail: rmcong@tju.edu.cn; jjlei@tju.edu.cn; hcp@tju.edu.cn).}
\thanks{H. Fu is with the Ocular Imaging Department, Institute for Infocomm Research, Agency for Science, Technology and Research, Singapore (e-mail: huazhufu@gmail.com).}
\thanks{Q. Huang is with the School of Computer and Control Engineering, University of Chinese Academy of Sciences, Beijing 100190, China (e-mail: qmhuang@ucas.ac.cn).}
\thanks{X. Cao is with State Key Laboratory of Information Security, Institute of Information Engineering, Chinese Academy of Sciences, Beijing 100093, China, and also with University of Chinese Academy of Sciences (e-mail: caoxiaochun@iie.ac.cn).}}

\markboth{IEEE TRANSACTIONS ON IMAGE PROCESSING, ~Vol.~xx, No.~xx, xxxx~2017}%
{Shell \MakeLowercase{\textit{et al.}}: Bare Demo of IEEEtran.cls for IEEE Journals}

\maketitle

\begin{abstract}
Co-saliency detection aims at extracting the common salient regions from an image group containing two or more relevant images. It is a newly emerging topic in computer vision community. Different from the most existing co-saliency methods focusing on RGB images, this paper proposes a novel co-saliency detection model for RGBD images, which utilizes the depth information to enhance identification of co-saliency. First, the intra saliency map for each image is generated by the single image saliency model, while the inter saliency map is calculated based on the multi-constraint feature matching, which represents the constraint relationship among multiple images. Then, the optimization scheme, namely Cross Label Propagation (CLP), is used to refine the intra and inter saliency maps in a cross way. Finally, all the original and optimized saliency maps are integrated to generate the final co-saliency result. The proposed method introduces the depth information and multi-constraint feature matching to improve the performance of co-saliency detection. Moreover, the proposed method can effectively exploit any existing single image saliency model to work well in co-saliency scenarios. Experiments on two RGBD co-saliency datasets demonstrate the effectiveness of our proposed model.
\end{abstract}


\IEEEpeerreviewmaketitle


\section{Introduction}

\IEEEPARstart{S}{ALIENCY} detection is a promising research area, which is regarded as a preferential allocation of computational resources \cite{R1,R2,R3,R3-2,R4}. It has been applied to a wide range of visual tasks, such as image retrieval \cite{R5}, image compression \cite{R6}, enhancement \cite{R7,R7-2,R7-3}, foreground annotation \cite{R8}, image retargeting \cite{R9,R10}, and pedestrian detection \cite{R10-2}. In recent years, co-saliency detection has become an emerging issue in saliency detection, which detects the common salient regions among multiple images \cite{R11,R12,R13,R14}. Different from the traditional single saliency detection model, co-saliency detection model aims at discovering the common salient objects from an image group containing two or more relevant images, while the categories, intrinsic characteristics, and locations of the salient objects are entirely unknown \cite{R15}. The co-salient objects simultaneously exhibit two properties, \emph{i.e.} 1) the co-salient regions should be salient with respect to the background in each image, and 2) all these co-salient regions should be similar in appearance among multiple images. Due to its superior expansibility, co-saliency detection has been widely used in many computer vision tasks, such as foreground co-segmentation \cite{R16,R17}, object co-localization and detection \cite{R18,R19,R20}, and image matching \cite{R21}.\par

Most existing co-saliency detection models are focused on RGB image and have achieved satisfactory performances \cite{R22,R23,R24,R25,R26,R27,R28,R29,R30}. However, little work has been done on co-saliency detection for RGBD images. Depth information has demonstrated its usefulness for many computer vision tasks, such as recognition \cite{R31}, object segmentation \cite{R32}, and saliency detection \cite{R33,R34,R35,R36,R37}. It reduces the ambiguity with color descriptors and enhances the identification of the object from the complex background. In this paper, the depth information is introduced as a novel cue for the co-saliency detection model. \par

For co-saliency detection, it is critical to effectively capture the consistent correspondence among multiple images. Fu \emph{et al.} \cite{R14} computed the inter saliency using a cluster-based method that integrates the multiple saliency cues. Cao \emph{et al.} \cite{R25} applied rank-one constraint to exploit the relationship of multiple saliency cues, and used the self-adaptive weight to generate the final co-saliency map. Cheng \emph{et al.} \cite{R28} proposed a group saliency model to extract the co-salient regions in a set of images via maximizing between-image similarities and within-image distinctness. Zhang \emph{et al.} \cite{R29} integrated the inter saliency calculation into a Bayesian framework. By contrast, in this paper, a multi-constraint feature matching algorithm at superpixel level is proposed to explore the inter image relationship, which integrates similarity constraint, saliency consistency, and cluster-based constraint. Introducing the multiple constraints into feature matching, the inter-image relationship will become more stable and robust.\par
In our work, we focus on solving two main issues: 1) how to effectively capture the corresponding relationship among multiple images, and 2) introduce the depth cue into co-saliency detection. To address these issues, a novel co-saliency model for RGBD images is proposed, which integrates the depth cue to enhance the identification of co-saliency. The multi-constraint based feature matching method is designed to capture the corresponding relationship and constrain the inter saliency map generation. Additionally, a Cross Label Propagation (CLP) method is proposed to optimize the intra and inter saliency maps in a cross way. The major contributions of the proposed co-saliency detection method are summarized as follows.\par
\begin{enumerate}[(1)]
\item To the best of our knowledge, our method is the first model that detects the co-salient objects from RGBD images. The depth information is demonstrated to be served as a useful complement for co-saliency detection.
\item A multi-constraint feature matching method is introduced to constrain the inter saliency map generation, which is robust to the complex backgrounds.
\item The CLP scheme is proposed to optimize the co-saliency model in our method.
\item The proposed method can effectively exploit any existing single image saliency model to work well in co-saliency scenarios.
\end{enumerate}\par
The rest of this paper is organized as follows. Section II reviews the related works of saliency and co-saliency detection. Section III introduces the proposed method in detail. The experimental results with qualitative and quantitative evaluations are presented in Section IV. Finally, the conclusion is drawn in Section V.\par

\section{Related Works}

In this section, we review the related works including the single saliency detection model and the co-saliency detection model.\par

\subsection{Single Saliency Detection Model}

Saliency detection model aims at highlighting the salient regions which can catch the human visual attention in the single image. Many saliency detection models for RGB image have been provided \cite{R38,R39,R40,R41,R42,R43,R44,R45}. Zhou \emph{et al.} \cite{R38} found the complementary relationship between compactness prior and local contrast, and designed a saliency detection model that integrates compactness and local contrast. Shi \emph{et al.} \cite{R40} proposed a hierarchical saliency detection model, which computes the saliency cues using weighted color contrast on three image layers. In \cite{R43}, a superpixel-wise convolutional neural network is proposed to learn the internal representations of saliency in an efficient manner. To address the blurry boundary of the salient object, Li \emph{et al.} \cite{R45} designed an end-to-end deep contrast network, which consists of two complementary components, namely pixel-level fully convolutional stream and segment-wise spatial pooling stream.\par

In addition, depth cue is also explored in the RGBD saliency detection models. Ju \emph{et al.} \cite{R33} proposed a depth-aware saliency method by using an anisotropic center-surround difference (ACSD) measure. Peng \emph{et al.} \cite{R34} proposed a multi-stage RGBD saliency detection model, which combines the low-level feature contrast, mid-level region grouping, and high-level prior enhancement. Feng \emph{et al.} \cite{R35} proposed an effective RGBD saliency feature using local background enclosure (LBE). Cong \emph{et al.} \cite{R36} computed the saliency for RGBD images using depth confidence analysis and multiple cues fusion. In \cite{R37}, a salient object detection method for RGBD images based on evolution strategy is proposed, which utilizes cellular automata to iteratively propagate the initial saliency map and generate the final detection result. Guo \emph{et al.} \cite{R54} proposed a RGBD saliency detection method based on saliency fusion and propagation. However, these methods only perform on single RGBD images, and can not be well applied in co-saliency models.\par
\begin{figure*}[!t]
\centering
\includegraphics[width=1\linewidth]{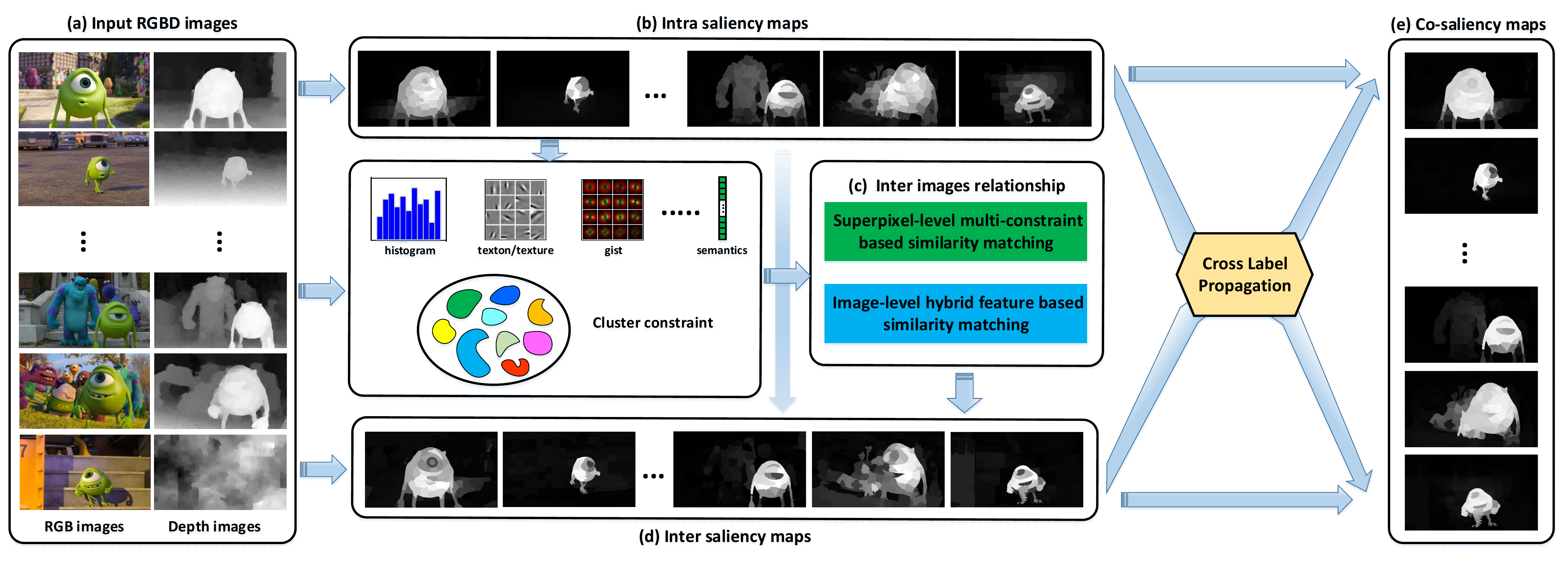}
\caption{The flowchart of the proposed RGBD co-saliency detection model. (a) The input RGBD images. (b) The intra saliency maps produced by existing single saliency method collaborating with depth information. (c) The inter-image corresponding relationship is obtained by superpixel-level multi-constraint based similarity matching and image-level hybrid feature based similarity matching. Using these corresponding relationships and intra saliency maps, the inter saliency maps (d) are generated. At last, the final co-saliency result (e) is achieved based on CLP. }
\label{fig1}
\end{figure*}
\subsection{Co-saliency Detection Model}

The goal of co-saliency detection model is to discover the common salient object from multiple images in an image group. Thus, capturing the constraint relationship among images is the key point for the co-saliency detection. The strategies for co-saliency detection can be grouped into three categories: bottom-up method, fusion-based method, and learning-based method. Li and Ngan \cite{R12} proposed a method to discover the salient objects via the combination of single saliency maps and an inter saliency map that is based on a co-multilayer graph. However, this method is only available for image pairs. Liu \emph{et al.} \cite{R22} proposed a hierarchical segmentation based co-saliency model, where the regional contrast based intra saliency map on the fine segmentation and object prior on the coarse segmentation are calculated respectively. In \cite{R23}, co-saliency detection is casted as a two-stage saliency propagation problem including inter-saliency propagation and intra-saliency propagation, and a new fusion strategy is used to obtain the final co-saliency result. In \cite{R46}, the multiscale segmentation voting based intra saliency map is integrated with image pairwise matching based inter saliency map for co-saliency detection.\par

Fusion-based method focuses on mining the useful knowledge from the existing saliency methods to generate the final co-saliency result. In \cite{R26}, a two-stage scheme led by queries is designed to obtain the guided saliency maps through a ranking mechanism, and the guided saliency maps are fused to generate the final co-saliency map. Huang \emph{et al.} \cite{R27} proposed a multiscale low-rank fusion method to detect the salient objects, and introduced a GMM-based co-saliency prior to discover the co-salient objects from multiple images.\par

In addition, learning-based method is proposed to handle the co-saliency detection as a classification problem. Zhang \emph{et al.} \cite{R29} proposed a unified co-salient object detection method via Bayesian framework. In the method, the high-level features are explored by using convolutional neural network (CNN) with additional transfer layers, and the visually similar neighbors from other image groups are introduced to suppress the common background regions. In \cite{R30}, the co-saliency detection is reformulated as a multiple-instance learning (MIL) problem and integrated into a self-paced learning (SPL) regime.\par

\section{Proposed Method}

In this section, we propose a co-saliency detection model for RGBD images. The flowchart of the proposed model is shown in Fig. \ref{fig1}. Firstly, the basic intra saliency map is generated by single saliency detection method associating with the depth cue on the individual image. Then, two similarity matching methods on different scales are designed to acquire the corresponding relationship among the multiple images. Combining the corresponding relationship and intra saliency map, the inter saliency map of each image is generated. In order to further improve the consistency of salient objects and suppress the background regions, a CLP optimization scheme is designed to refine the intra and inter saliency maps in a cross way. At last, the weighted averaging fusion is used to produce the final co-saliency result.\par

\subsection{Intra Saliency Detection}

Given $N$ input RGB images $\{I^{i}\}_{i=1}^{N}$ and corresponding depth maps $\{D^{i}\}_{i=1}^{N}$. We firstly employ SLIC algorithm \cite{R47} to abstract each RGB image $I^{i}$ into superpixels $R^{i}=\{r_{m}^{i}\}_{m=1}^{N_{i}}$, where $N_{i}$ denotes the number of superpixels. Then, our previous work in RGBD saliency detection \cite{R36} is exploited to generate the intra saliency map for each RGBD images. The intra saliency value of a superpixel $r_{m}^{i}$ in image $I^{i}$, namely $S_{intra}(r_{m}^{i})$, is assigned with the mean value of all pixels that belong to superpixel $r_{m}^{i}$ in the intra saliency map.\par
Note that, any single saliency method can be utilized to generate the intra saliency map. Here, we choose the work in \cite{R36} due to its high effectiveness and robustness for RGBD saliency detection. Generally, the more accurate the intra saliency map is, the better it is for the co-saliency computation using our model. The experimental comparisons of different intra saliency maps are discussed in Section IV-E.\par

\subsection{Inter Saliency Detection}
Different from the single saliency detection model, co-saliency detection aims at discovering the common salient regions from an image group. Accordingly, acquiring the corresponding relationship among multiple images is the key point of co-saliency detection model. In the proposed model, the matching methods on two levels are designed to represent the correspondence among multiple images. The first one is the superpixel-level similarity matching scheme, which focuses on determining the matching superpixel set for the current superpixel based on three constraints from other images . The second is the image-level similarity measurement, which provides a global relationship on the whole image scale. With the corresponding relationship, the inter saliency of a superpixel is defined as the weighted sum of the intra saliency of corresponding superpixels in other images.\par

\subsubsection{Superpixel-level multi-constraint based similarity matching}

At the superpixel level, the correspondence is represented as the multi-constraint based matching relationship between the superpixels among the multiple images, which considers the similarity constraint, saliency consistency, and cluster-based constraint.\par

\emph{\textbf{Similarity constraint.}} In our work, the color and depth cues are simultaneously considered to express the similarity constraint. However, for some RGBD images, the depth map is seriously noisy, which may degenerate the accuracy of the measurement. To address this issue, the depth confidence measure $\lambda_{d}$ is introduced to evaluate the reliability of depth map as used in \cite{R36}.
\begin{equation}
\lambda_{d}=\exp((1-m_{d})\cdot\emph{CV}\cdot\emph{H})-1
\end{equation}
where $m_{d}$ is the mean value of the whole depth map, $\emph{CV}=m_{d}/\sigma_{d}$ is the coefficient of variation, $\sigma_{d}$ is the standard deviation for depth image,  $\emph{H}=-\sum_{i=1}^L(n_{i}/n_{\Sigma})\log(n_{i}/n_{\Sigma})$ is the depth frequency entropy, which denotes the randomness of depth distribution, $n_{\Sigma}$ is the number of pixels in the depth map, $n_{i}$ is the number of pixels that belong to the region level $rl_{i}$, and $L$ is the levels of depth map. A larger $\lambda_{d}$ corresponds to more reliable of the input depth map. Thus, $\lambda_{d}$ is used as a controller for the introduction of depth information in our model. Then, the similarity matrix $\textbf{\emph{S}}=[s(r_{m}^{i},r_{n}^{j})]_{N_{i}\times N_{j}}$ between superpixels from the $i^{th}$ and $j^{th}$ images is defined as:
\begin{equation}
s(r_{m}^{i},r_{n}^{j})=\exp(-\frac{\|\textbf{\emph{c}}_{m}^{i}-\textbf{\emph{c}}_{n}^{j}\|_{2}+\min(\lambda_{d}^{i},\lambda_{d}^{j})\cdot{|d_{m}^{i}-d_{n}^{j}|}}{\sigma^{2}})
\end{equation}
where $\textbf{\emph{c}}_{m}^{i}$ is the mean color vector of superpixel $r_{m}^{i}$ in the L*a*b* color space, $d_{m}^{i}$ denotes the mean depth value of superpixel $r_{m}^{i}$, $\lambda_{d}^{i}$ represents the depth confidence measure of depth map $D^{i}$, $\|\cdot\|_{2}$ denotes the $\ell_{2}$-norm, and $\sigma^{2}$ is a parameter to control strength of the similarity, which is fixed to 0.1. Based on the similarity matrix in Eq. (2), the $K_{max}$ nearest neighbors in each of other images for superpixel $r_{m}^{i}$ can be determined. Then, these neighbor superpixels for $r_{m}^{i}$ are composed as the set $\Phi_{1}(r_{m}^{i})$.\par

\emph{\textbf{Saliency consistency.}} Considering the task of co-saliency detection, the saliency consistency is introduced as another important cue to constrain the feature matching. Thus, we calculate the saliency similarity between the target superpixel $r_{m}^{i}$ and other superpixels $\{r_{n}^{j}\}_{n=1}^{N_{j}}$, and select the superpixels with consistent saliency values to generate the set for $r_{m}^{i}$ as:
\begin{equation}
\Phi_{2}(r_{m}^{i})=\{r_{n}^{j}||S_{intra}(r_{m}^{i})-S_{intra}(r_{n}^{j})|\leq T_{1}\}
\end{equation}
where $n=\{1,2,\ldots,N_{j}\}$, and $T_{1}$ is a threshold to control strength of the saliency similarity, which is fixed to 0.3.\par

\emph{\textbf{Cluster-based constraint.}} Inspired by the fact that the matching superpixels should be grouped into the same cluster, the cluster-based constraint is introduced to build the cluster-level correspondence. At first, \emph{K-means++} clustering \cite{R48} is used to group the superpixels $\{r_{m}^{i}\}_{m=1}^{N_{i}}$ into $K$ clusters $\{C_{k}^{i}\}_{k=1}^{K}$ with cluster centers $\{c_{k}^{i}\}_{k=1}^{K}$. Then, the Euclidean distance is utilized to measure and determine the cluster-level superpixel matching relationship. For each superpixel $r_{m}^{i}$, we can find one superpixel with the minimum distance in each of other images. Supposed that the superpixel $r_{m}^{i}$ belongs to the cluster $C_{p}^{i}$, and the superpixel $r_{n}^{j}$ belongs to the cluster $C_{q}^{j}$. The cluster-level nearest neighbor superpixels for superpixel $r_{m}^{i}$ are denoted as $\Phi_{3}(r_{m}^{i})$.
\begin{equation}
\Phi_{3}(r_{m}^{i})=\{r_{n}^{j}| \arg{\min_{C_{q}^{j},q\in[1,K]}{Ed(c_{p}^{i},c_{q}^{j})}}\}
\end{equation}
where $Ed$ denotes the Euclidean distance function, $c_{p}^{i}$ and $c_{q}^{j}$ are the cluster centers of cluster $C_{p}^{i}$ and $C_{q}^{j}$, respectively.\par
\emph{\textbf{Similarity matching.}} Three corresponding sets $\Phi_{1}$, $\Phi_{2}$, and $\Phi_{3}$ are combined to determine the matching relationship for each superpixel. The matching matrix $\textbf{\emph{ML}}^{ij}=[ml(r_{m}^{i},r_{n}^{j})]_{N_{i}\times N_{j}}$ is defined as:
\begin{equation}
ml(r_{m}^{i},r_{n}^{j})=
\begin{cases}
1, & \text{if $r_{n}^{j}\in\{\Phi_{1}(r_{m}^{i})\cap\Phi_{2}(r_{m}^{i})\cap\Phi_{3}(r_{m}^{i})\}$}\\
0, & \text{otherwise}
\end{cases}
\end{equation}

\subsubsection{Image-level hybrid feature based similarity matching}
\begin{table}[!t]
\renewcommand\arraystretch{1.8}
\centering
\caption{the Image Property Descriptor and the Feature Distance.}
\begin{tabular}{c|c|c|c||c}
\toprule[1.5pt]
  & features & description & dim & distance\\[0.5ex]
\hline
\hline
\multirow{4}{*}{col}&$\textbf{\emph{h}}_{c}$&RGB histogram&512&$d_{c1}=\chi^{2}(\textbf{\emph{h}}_{c}^{i},\textbf{\emph{h}}_{c}^{j})$\\
  &$\textbf{\emph{t}}$&texton histogram&15&$d_{c2}=\chi^{2}(\textbf{\emph{t}}^{i},\textbf{\emph{t}}^{j})$\\
  &$\textbf{\emph{s}}$&semantic feature&4096&$d_{c3}=1-\cos{(\textbf{\emph{s}}^{i},\textbf{\emph{s}}^{j})}$\\
  &$\textbf{\emph{g}}$&GIST feature&512&$d_{c4}=1-\cos{(\textbf{\emph{g}}^{i},\textbf{\emph{g}}^{j})}$\\
\hline
dep&$\textbf{\emph{h}}_{d}$&depth histogram&512&$d_{d}=\chi^{2}(\textbf{\emph{h}}_{d}^{i},\textbf{\emph{h}}_{d}^{j})$\\
\hline
sal&$\textbf{\emph{h}}_{s}$&saliency histogram&512&$d_{s}=\chi^{2}(\textbf{\emph{h}}_{s}^{i},\textbf{\emph{h}}_{s}^{j})$\\
\bottomrule[1.5pt]
\end{tabular}
\label{tab1}
\end{table}
Enlightened by the observation that the greater similarity between two images means the greater likelihood of finding the matching regions, a full-image size similarity descriptor is designed as the  weighted coefficient for inter saliency calculation. To evaluate the image similarity, three types of features are used to represent the image property from different aspects and guarantee the completeness and generality of feature selection. Firstly, the color feature, as the common basic feature in most saliency detection methods, is introduced in our method. Inspired by the fact that similar images should have approximate depth distributions and similar appearances in salient objects, the depth and saliency histograms are added in the feature pool. At last, these feature distances are integrated through a self-adaptive weighted strategy to evaluate the similarity between two images.\par

The features used in our method are listed in TABLE \ref{tab1}. The details of the features are described as follows: the color histogram in the RGB color space is utilized to represent the color distribution; the texton histogram is used to express the texture feature; and the GIST feature \cite{R49} is introduced to describe the spatial structure of the scene. In addition, the deep feature produced by VGG network \cite{R50} is used to represent the semantic information of the image. The fc7 feature with pre-trained VGG16 model on ImageNet is directly extracted as the semantic feature in our method. Moreover, the depth and saliency histograms are used to describe the distributions of the depth map and single saliency map.\par
Then, the feature distances between two images are summarized in the last column of TABLE \ref{tab1}, where $\chi^{2}(\cdot)$ represents the Chi-square distance, and $\cos(\cdot)$ denotes the cosine distance. Finally, these feature distances are fused to evaluate the image similarity as:
\begin{equation}
\varphi^{ij}=1-(\alpha_{c}\cdot\Sigma_{i=1}^{4}{d_{ci}/4}+\alpha_{d}\cdot d_{d}+\alpha_{s}\cdot d_{s})
\end{equation}
where $\varphi^{ij}$ denotes the similarity measurement between the $i^{th}$ and $j^{th}$ images, $\alpha_{c}$, $\alpha_{d}$, and $\alpha_{s}$ are the coefficients for color, depth, and saliency feature distances, respectively. A larger $\varphi^{ij}$ corresponds to higher similarity between the two input images. The coefficients are set based on three criteria: 1) The sum of coefficients should be 1, as $\alpha_{c}+\alpha_{d}+\alpha_{s}=1$. 2) The color and saliency distances are assigned the same weight for simplicity. 3) The poor depth map, like a noise, may have a negative influence on the measurement. Therefore, a self-adaptive weighted coefficient for depth distance is designed according to the depth confidence measure $\lambda_{d}$.
\begin{equation}
\alpha_{d}=
\begin{cases}
\lambda_{d}^{min}, & \text{if $\lambda_{d}^{min}=\min{(\lambda_{d}^{i},\lambda_{d}^{j})}\leq T_{2}$}\\
1/3, & \text{otherwise}
\end{cases}
\end{equation}
and
\begin{equation}
\alpha_{c}=\alpha_{s}=\frac{1}{2}\cdot(1-\alpha_{d})
\end{equation}
where $T_{2}$ is a threshold to distinguish the degenerated depth map, and it is set to 0.2 in the experiments.\par

\subsubsection{Inter saliency calculation}
After obtaining the corresponding relationship among multiple images through the superpixel-level feature matching and image-level similarity matching, the inter saliency of a superpixel is computed as the weighted sum of the intra saliency of corresponding superpixels in other images. The superpixel-level feature matching result provides the corresponding relationship between the superpixels among different images, and the weighted coefficient is calculated by the image-level similarity measurement.\par
With the matching matrix $\textbf{\emph{ML}}^{ij}$, image similarity $\varphi^{ij}$, and intra saliency map $S_{intra}$, the inter saliency value of each superpixel is assigned as:
\begin{equation}
S_{inter}(r_{m}^{i})=\frac{1}{N-1}\sum_{j=1,j\neq i}^{N}\frac{\varphi^{ij}}{N_{j}}\sum_{n=1}^{N_{j}} S_{intra}(r_{n}^{j})\cdot ml(r_{m}^{i},r_{n}^{j})
\end{equation}
where $r_{m}^{i}$ denotes the $m^{th}$ superpixel in image $I^{i}$, $N$ represents the number of images in the group, $N_{j}$ is the number of superpixels in the $j^{th}$ image, and $\varphi^{ij}$ is the similarity measurement between the $i^{th}$ and $j^{th}$ images.\par
\begin{figure}[!t]
\centering
\includegraphics[width=1\linewidth]{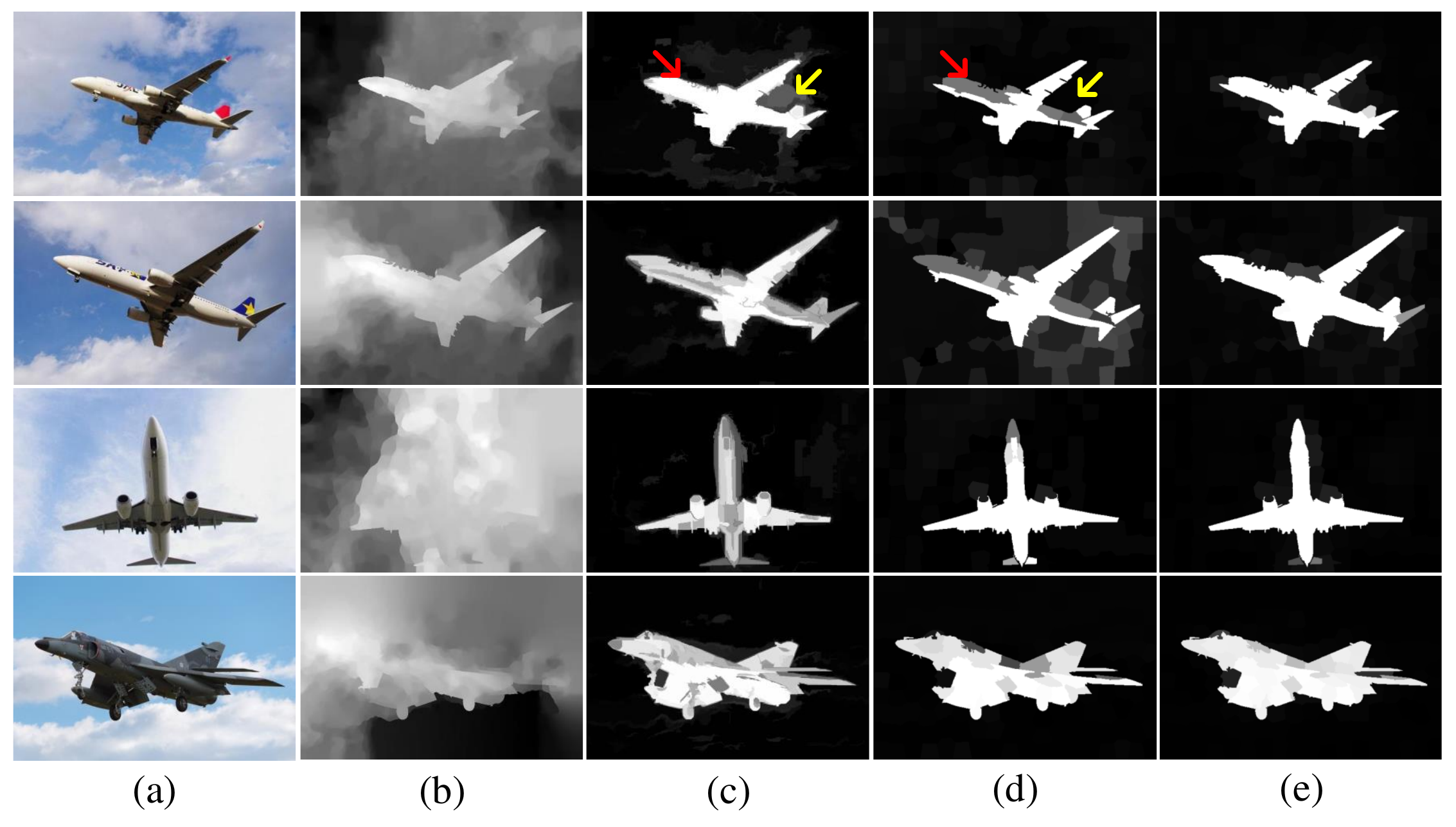}
\caption{Some examples of our proposed method. (a) The RGB image. (b) The depth map. (c) The intra saliency map. (d) The inter saliency map. (e) The final co-saliency results with CLP optimization.}
\label{fig2}
\end{figure}

Some visual examples of the proposed method are shown in Fig. \ref{fig2}. The third column represents the intra saliency maps, and the fourth column shows the inter saliency maps. From the figure, we can see that the intra and inter saliency maps provide different useful information. For example, at the first row of Fig. \ref{fig2}, some common backgrounds in the intra saliency map (\emph{e.g.} sky marked by the yellow arrow) are wrongly detected as the salient regions, while are effectively suppressed in the inter saliency map. In addition, the fuselage marked by the red arrow is detected homogeneously in the intra saliency map, but failed in the inter saliency map. Therefore, in order to generate more consistent and accurate co-saliency map, the optimization and fusion schemes are proposed in the following sections.\par

\subsection{Optimization and Propagation}

In the proposed method, the optimization of saliency map is casted as a ``label propagation'' problem, where the uncertain labels are propagated by using two types of certain seeds, \emph{i.e.} background and salient seeds. The proposed CLP method is used to optimize the intra and inter saliency maps in a cross way, which means the propagative seeds are crosswise interacted. The cross seeding strategy optimizes the intra and inter saliency maps jointly, and improves the robustness.\par

In the CLP optimization algorithm, the graph $G=(V,E)$ is firstly constructed, where $V$ represents the set of nodes which corresponds to the superpixels, and $E$ denotes the set of links between adjacent nodes. The affinity matrix $\textbf{\emph{W}}^{i}=[w_{uv}^{i}]_{N_{i}\times N_{i}}$ is defined as the similarity between two adjacent superpixels in the $i^{th}$ image:
\begin{equation}
w_{uv}^{i}=
\begin{cases}
\exp(-\frac{\|\textbf{\emph{c}}_{u}^{i}-\textbf{\emph{c}}_{v}^{i}\|_{2}+\lambda_{d}^{i}\cdot{|d_{u}^{i}-d_{v}^{i}|}}{\sigma^{2}}), & \text{if $v\in\Psi_{u}^{i}$}\\
0, & \text{otherwise}
\end{cases}
\end{equation}
where $\Psi_{u}^{i}$ is the set of neighbors of superpixel $r_{u}^{i}$.\par
Taking the optimization of intra saliency map as an example, the detailed procedures are described as follows. The certain seeds, including foreground labeled seeds $F$ and background labeled seeds $B$, are selected to update and optimize the saliency of unlabeled nodes $U$. We design two thresholds to determine these labeled seeds.
\begin{equation}
TF(S)=\max(\frac{2}{N_{i}}\sum_{m=1}^{N_{i}}|S(r_{m}^{i})|,T_{min})
\end{equation}
\begin{equation}
TB(S)=\min(\frac{1}{N_{i}}\sum_{m=1}^{N_{i}}|S(r_{m}^{i})|,T_{max})
\end{equation}
where $S(r_{m}^{i})$ denotes the intra or inter saliency score of the superpixel $r_{m}^{i}$, $TF(S)$ is a threshold of saliency map $S$ for foreground seeds selection, $T_{min}$ is the minimum threshold for $TF(S)$, $TB(S)$ is a threshold for background seeds selection, and $T_{max}$ is the maximum threshold for $TB(S)$. Then, these thresholds are used to determine the set of labeled seeds and initialize the saliency score of superpixels. The saliency scores of superpixels in CLP method are initialized as follows.
\begin{equation}
V_{0}^{CLP}(r_{m}^{i})=
\begin{cases}
1, & \text{if $S_{inter}(r_{m}^{i})\geq TF(S_{inter})$}\\
0, & \text{if $S_{inter}(r_{m}^{i})\leq TB(S_{inter})$}\\
S_{intra}(r_{m}^{i}), & \text{otherwise}
\end{cases}
\end{equation}\par
Once the initialization is completed, the saliency is propagated using the labeled seeds on the graph according to the equation:
\begin{equation}
V_{intra}^{CLP}(r_{m}^{i})=\sum_{n=1}^{N_{i}}w_{mn}^{i}V_{0}^{CLP}(r_{n}^{i})
\end{equation}
After normalization, the optimized intra saliency map $S_{intra}^{CLP}=norm(V_{intra}^{CLP})$ is achieved, where $norm(x)$ is a function that normalizes $x$ to [0,1] using min-max normalization. The optimized inter saliency map $S_{inter}^{CLP}$ is generated using the same procedure conducted on the inter saliency map. It should be noted that the inter saliency map is firstly optimized using the intra saliency map in CLP method. Then, the optimized inter saliency map is used to update the intra saliency map, since the inter saliency is generated by the intra saliency map. In order to guarantee the optimization performance, the inter saliency map is firstly optimized in our model.\par
\begin{figure*}[!t]
\centering
\includegraphics[width=1\linewidth]{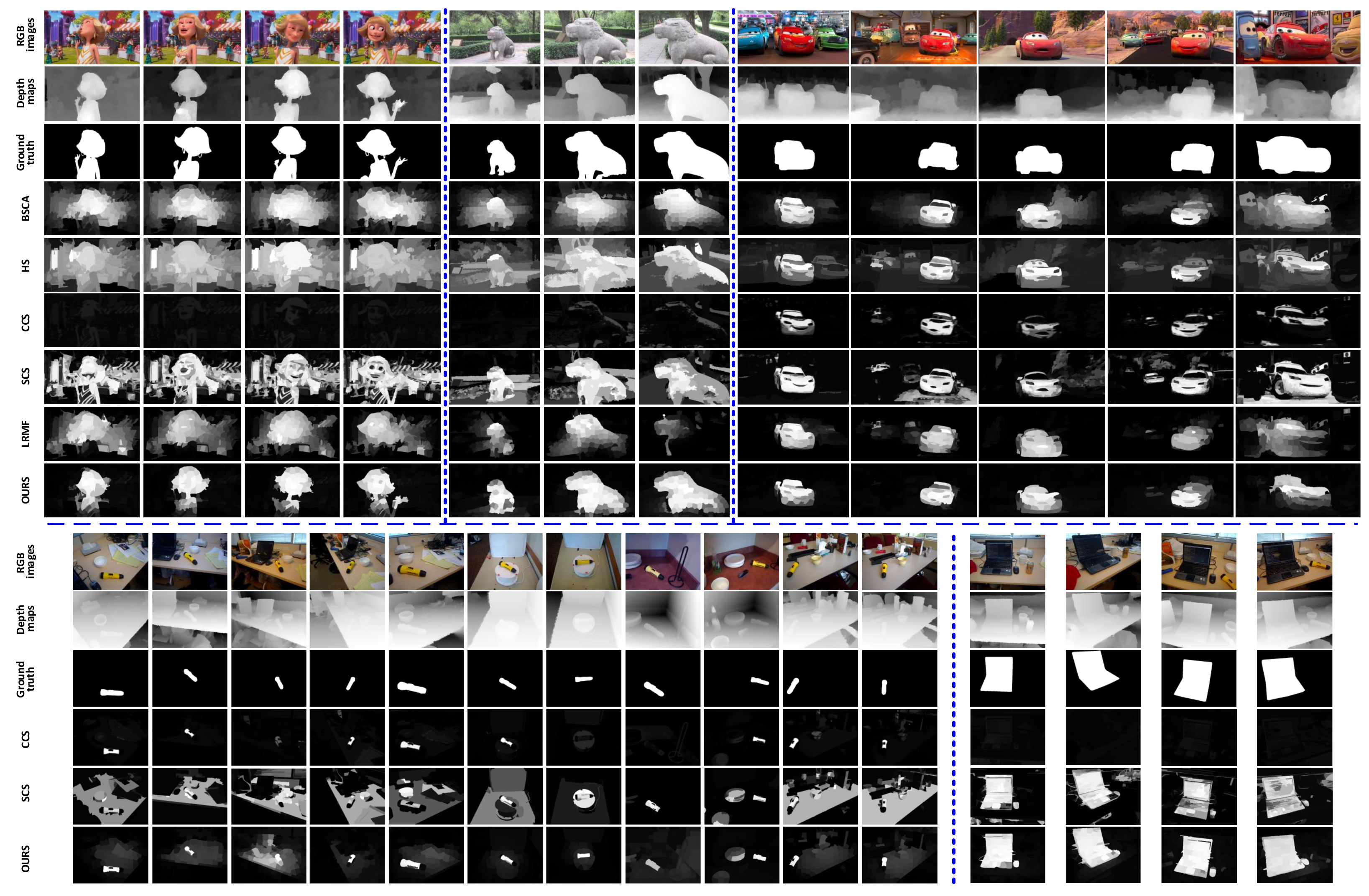}
\caption{Visual examples of different saliency and co-saliency detection methods on two datasets. }
\label{fig3}
\end{figure*}
\subsection{Co-saliency Detection}
Finally, the initial intra/inter saliency maps and the optimized intra/inter saliency maps are integrated to generate the final co-saliency map.
\begin{equation}
S_{co}^{CLP}=\gamma_{1}\cdot S_{intra}+\gamma_{2}\cdot S_{inter}+\gamma_{3}\cdot S_{intra}^{CLP}+\gamma_{4}\cdot S_{inter}^{CLP}
\end{equation}
where $\gamma_{i}$ is the weighted coefficient with $\Sigma_{i=1}^{4}\gamma_{i}=1$. Without loss of generality, these four parameters are all set to 0.25 in experiments. The last column of Fig. \ref{fig2} shows the final co-saliency results. In terms of consistency and accuracy, the final co-saliency result is significantly improved after optimization and fusion.

\section{Experiments}
In this section, we first introduce the experimental settings, which include the datasets, implementation details, and evaluation metrics. Then, the qualitative and quantitative comparisons are presented in Section IV-B. The evaluation of different parameters are analyzed in Sections IV-C, IV-D, and IV-E. At last, some degenerated cases are discussed in Section IV-F.\par

\subsection{Experimental Settings}
\subsubsection{Datasets and implementation details}
We evaluate the proposed co-saliency model on two datasets. The RGBD Coseg183 dataset \cite{R17} is a co-segmentation dataset which contains 183 images in total that are distributed in 16 image sets, and each of 6 to 17 images is taken from indoor scenes with one common foreground object. Pixel-level ground-truth is manually labeled for each image in the dataset\footnote{ \url{http://hzfu.github.io/proj_rgbdseg.html} }. It is a challenging dataset for co-saliency detection due to the multiple objects in each image. Additionally, we also construct a new RGBD co-saliency dataset, named RGBD Cosal150 dataset\footnote{  \url{https://rmcong.github.io/proj_RGBD_cosal.html} }. In this dataset, we collected 21 image groups containing totally 150 images from RGBD NJU-1985 dataset \cite{R33} with pixel-level ground truth, and the original depth maps are provided by the dataset itself.\par

In our method, two hundred superpixels are generated for each image using the SLIC method, \emph{i.e.} $N_{i}=200$. The thresholds for seeds selection are assigned to $T_{min}=0.6$ and $T_{max}=0.2$. Simultaneously, the number of clusters $K$ used in \emph{K-means++} is set to 10, and the maximum matching number $K_{max}$ for feature matching is set to 40. The proposed method is implemented in MATLAB 2014a, and all the experiments are performed on a Quad Core 3.5GHz workstation with 16GB RAM. Our algorithm costs average 41.03 seconds to process one image on the RGBD Cosal150 dataset. In the proposed method, the intra saliency calculation costs 27.86\% running time, the inter saliency model takes 71.94\% running time, and the optimization stage consumes 0.20\% running time.\par

\subsubsection{Evaluation metrics}
To evaluate the performance of the proposed method, four criteria including the Precision-Recall (PR) curve, the  Precision and Recall scores, F-measure, and Mean Absolute Error (MAE) are calculated. The precision and recall scores are produced by thresholding the saliency map into binary salient object masks with a series of fixed integers from 0 to 255. A weighted mean of precision and recall is employed to calculate the F-measure \cite{R51}, which is expressed as:
\begin{equation}
F_{\beta}=\frac{(1+\beta^{2})Precision\times Recall}{\beta^{2}\times Precision+ Recall}
\end{equation}
where we set $\beta^{2}=0.3$ to emphasize the precision more than recall. In addition, Mean Absolute Error (MAE) score directly measures the difference between the continuous saliency map $S$ and ground truth $G$ as:
\begin{equation}
MAE=\frac{1}{W\times H}\sum_{x=1}^{W}\sum_{y=1}^{H}|S(x,y)-G(x,y)|
\end{equation}
where $W$ and $H$ denote the width and height of the input image.\par

\subsection{Comparison with State-of-the-art Methods}
\begin{figure*}[!t]
\centering
\includegraphics[width=1\linewidth]{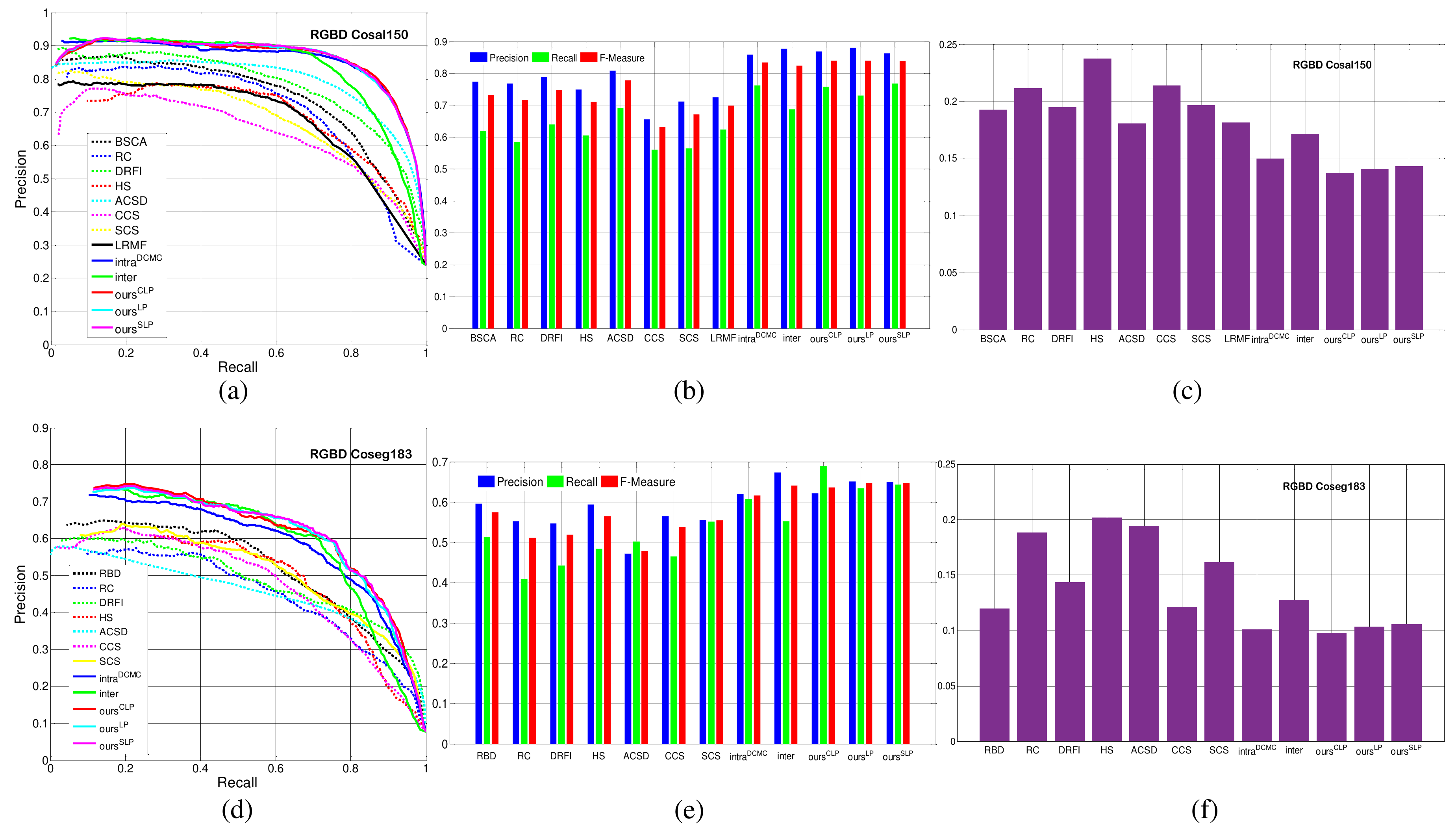}
\caption{The quantitative performances of different methods on two datasets. Notice that ``our*'' means implementing our method using different optimization approaches, where $*=\{\text{LP, SLP, CLP}\}$. (a)-(c) PR curves, Precision and Recall scores, F-measure, and MAE scores on the RGBD Cosal150 dataset. (d)-(f) PR curves, Precision and Recall scores, F-measure, and MAE scores on the RGBD Coseg183 dataset.}
\label{fig4}
\end{figure*}

We compare the proposed method with some state-of-the-art saliency/co-saliency detection methods, \emph{i.e.} HS \cite{R40}, BSCA \cite{R41}, RC \cite{R52}, DRFI \cite{R53}, ACSD \cite{R33}, DCMC \cite{R36}, CCS \cite{R14}, SCS \cite{R26}, and LRMF \cite{R27}, in which the first six methods are single saliency detection models and the last three ones are co-saliency methods for RGB image. In addition, other two optional optimization mechanisms, namely Label Propagation (LP) and Shared Label Propagation (SLP), are reported as the baselines in the experiments. The main difference of these three methods lies in the selection of certain seeds. The LP scheme determines the seeds from their own intra or inter saliency map. For the SLP mechanism, the intra and inter saliency maps are combined to determine the common seeds. Then, the selected seeds are shared to optimize the intra and inter saliency maps simultaneously. By contrast, the propagative seeds are crosswise interacted for the CLP scheme, and it bridges the gap between the intra saliency and inter saliency in the process of optimization. Therefore, in principle, the CLP scheme is more suitable for co-saliency detection due to the interactive information from intra and inter saliency maps.\par

Some visual comparisons of different methods on two datasets are illustrated in Fig. \ref{fig3}, which contain five image groups: \emph{woman}, \emph{sculpture}, \emph{car}, \emph{yellow flashlight}, and \emph{computer}. From the figure, even though the images own complex and variable backgrounds or the salient objects exhibit large variations in shape and direction, the proposed method effectively highlights the common salient objects from the image group. Furthermore, the results produced by our model are more accurate and uniform than other methods. For example, in the \emph{woman} group, the eyes and mouth of the woman are wrongly detected as background regions through the SCS model \cite{R26}, and some background regions are also detected as salient regions due to their complex textures. By contrast, the woman in different images are uniformly detected with clearly contour by the proposed method. The same situation is faced to the LRMF method \cite{R27}, where the body of the woman is missed and the background regions are wrongly detected.\par

The quantitative comparison results in terms of the PR curves, precision and recall scores, F-measure, and MAE scores are reported in Fig. \ref{fig4}. Before comparing with other methods, we first analyze the results of our method in different stages, which include intra and inter saliency modeling, and co-saliency generation with three different optimization schemes (\emph{i.e.} LP, SLP, and CLP ). It can be observed that 1) the inter saliency map performs a better result compared with other existing co-saliency methods, and 2) the performance of co-saliency result with the optimization model is obviously improved compared with the intra and inter saliency maps individually. Moreover, consistent with the theoretical analysis, the CLP scheme achieves favorable performance on the two datasets. For example, the proposed method with CLP optimization achieves the best performance on the RGBD Cosal150 dataset according to the comprehensive measures (\textrm{F-measure} = 0.8403, and \textrm{MAE} = 0.1370), and it also performs the best result on RGBD Coseg183 dataset in terms of MAE measure (\textrm{F-measure} = 0.6365, and \textrm{MAE} = 0.0979). For the SLP optimization strategy, its performance is slightly worse than other methods, where the MAE score is 0.1430 on the RGBD Cosal150 dataset, and 0.1052 on the RGBD Coseg183 dataset. The main reason is that the shared way for seed selection enables reduction in the number of seeds, which in turn degrades the propagation performance.\par

Compared with other single saliency and co-saliency methods, the proposed model achieves the highest precisions of the whole PR curves on both two datasets. In addition, the proposed co-saliency model achieves the best performance on both of the RGBD Coal150 and Coseg183 datasets with the highest F-measure and the smallest MAE score. For the F-measure, our co-saliency model achieves a maximum percentage gain of 30.67\% compared to other saliency results on the RGBD Cosal150 dataset, and the minimum percentage gain also reaches 5.88\%. The maximum and minimum percentage gains of the MAE score achieve 46.00\% and 24.14\%, respectively. On the RGBD Coseg183 dataset, the proposed method also obtains obvious performance gains. For example, the F-measure and MAE score are at least increased by 10.67\% and 18.21\%, respectively. \par

In summary, benefiting from the two-level similarity matching and CLP optimization, our co-saliency detection model achieves a satisfying performance. The visual comparisons and quantitative analyses demonstrate the effectiveness of the proposed model. We will discuss the influence of some parameters in the next subsections.\par
\subsection{Evaluation of the Maximum Matching Number $K_{max}$}
\begin{figure}[!b]
\centering
\includegraphics[width=7.6cm,height=5.8cm]{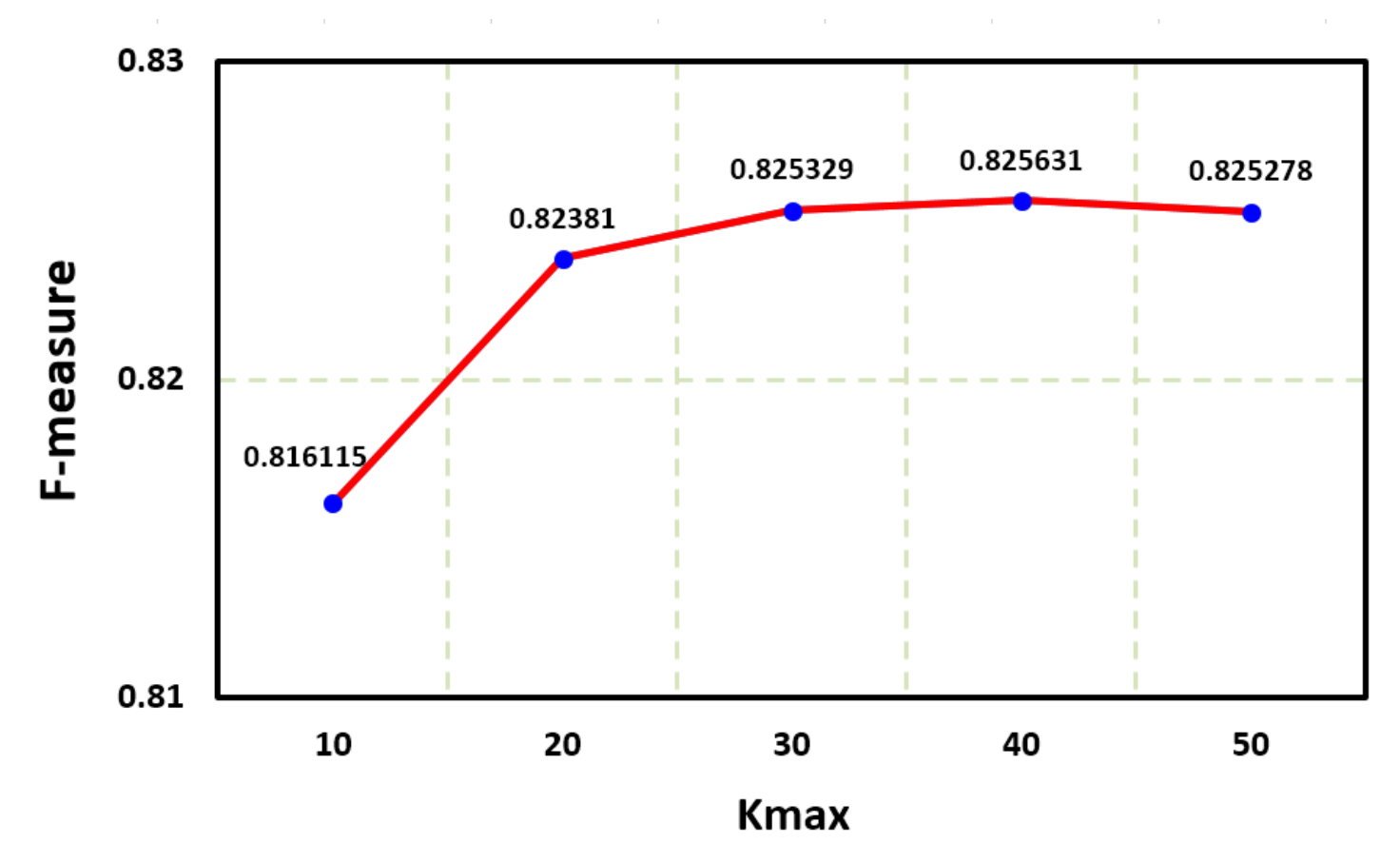}
\caption{The F-measure of inter saliency map on the RGBD Cosal150 dataset under different maximum matching numbers.}
\label{fig5}
\end{figure}
We conduct experiments on RGBD Cosal150 dataset to analyze the influence of the maximum matching number $K_{max}$ in the procedure of inter saliency calculation of the proposed method. Considering the parameter $K_{max}$ is only used to calculate the inter saliency map, the F-measures of the inter saliency maps are evaluated under different $K_{max}$, as shown in Fig. \ref{fig5}. From the curve of F-measure with different $K_{max}$, we can observe that the inter saliency maps with different maximum matching numbers achieve the comparable performance except for the result with $K_{max}=10$. The main reason is that $K_{max}=10$ is too small to obtain a relatively stable and accurate matching result. The performance will become stable when the $K_{max}$ reaches 30. Considering the number of superpixel in an image is set to 200 in our experiments, we set the maximum matching number $K_{max}$ to be 40 balancing the computational complexity and accuracy. In conclusion, the performance of inter saliency map is not highly sensitive to the parameter $K_{max}$. In general, due to the saliency consistency and cluster-based constraint are introduced in our model, the maximum matching number between superpixels among two images in the process of feature matching can be set to a larger value, such as the one-fifth of the superpixel number in an image.\par

\subsection{Evaluation of the Depth Cue}
\begin{figure}[!t]
\centering
\includegraphics[width=1\linewidth]{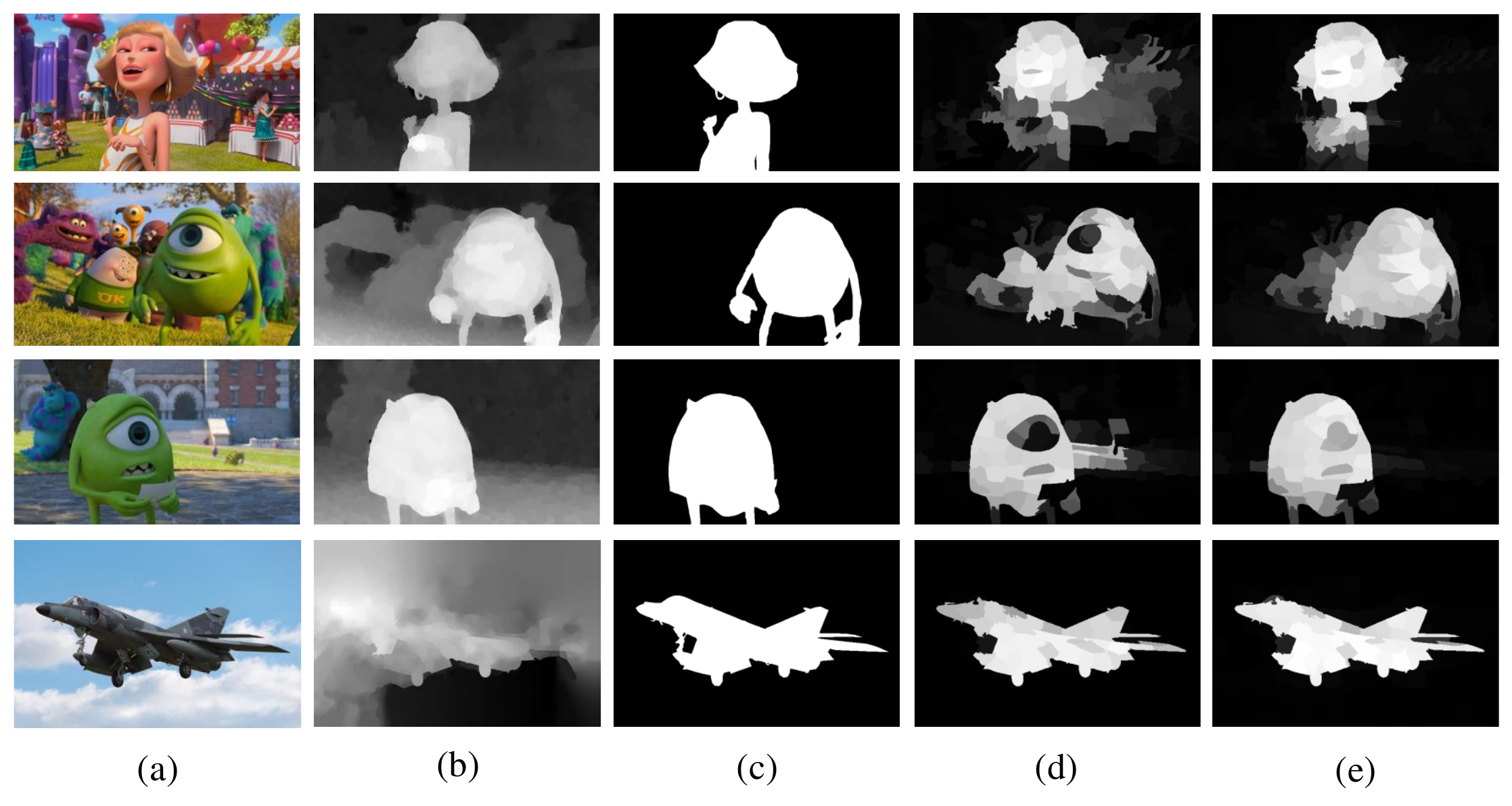}
\caption{The visual examples with and without depth information through the proposed co-saliency detection method. (a) RGB images. (b) Depth maps. (c) Ground truth. (d) Saliency maps without depth information. (e) Saliency maps with depth information.}
\label{fig6}
\end{figure}

In our model, the depth cue is introduced to assist the identification of co-salient regions. To evaluate the depth cue on the whole framework, we conduct an experiment to evaluate the effect of the depth cue on the RGBD Cosal150 dataset, and the results are shown in Fig. \ref{fig6} and TABLE \ref{tab3}. \par

For the depth map, in most cases, it is clear and effective, and exhibits great power for improving the saliency performance, such as the first three rows in Fig. \ref{fig6}. The depth information can be regarded as an effective cue to distinguish the foreground from the complex background. Therefore, utilizing the depth cue, the complex and cluttered background regions (such as the stores, other cartoons, and lawns) are suppressed obviously, and the salient objects are better highlighted. However, in some cases, the depth map has poor quality and may degrade the performance. To address this problem, the depth confidence measure is introduced as a weight to control the contribution of depth information. In this poor quality case, as shown in the last row of Fig. \ref{fig6}, even with the noised depth information, our model still achieved relatively satisfying performance similar to the RGB co-saliency model. From the quantitative measures reported in TABLE \ref{tab3}, without the depth cue, the F-measure of our model achieves 0.7639, and the MAE score reaches 0.1599. With the depth cue, the overall performance of F-measure is increased to 0.8403, and the MAE score is also improved to 0.1370. In summary, the depth information is a useful and effective cue for the co-saliency detection. \par

\begin{table}[!t]
\renewcommand\arraystretch{1.8}
\centering
\caption{Quantitative Evaluations with and without Depth Cue through the Proposed Co-saliency Detection Method on the RGBD Cosal150 Dataset}
\begin{tabular}{c|c|c|c}
\toprule[1.5pt]
  & without depth & with depth & percentage gain \\[0.5ex]
\hline
F-measure & $0.7639$ & $0.8403$ & $10.00\%$ \\
\hline
MAE & $0.1599$ & $0.1370$ & $14.32\%$\\
\bottomrule[1.5pt]
\end{tabular}
\label{tab3}
\end{table}
\subsection{Evaluation of the Different Intra Saliency Methods}
\begin{figure}[!b]
\centering
\includegraphics[width=1\linewidth]{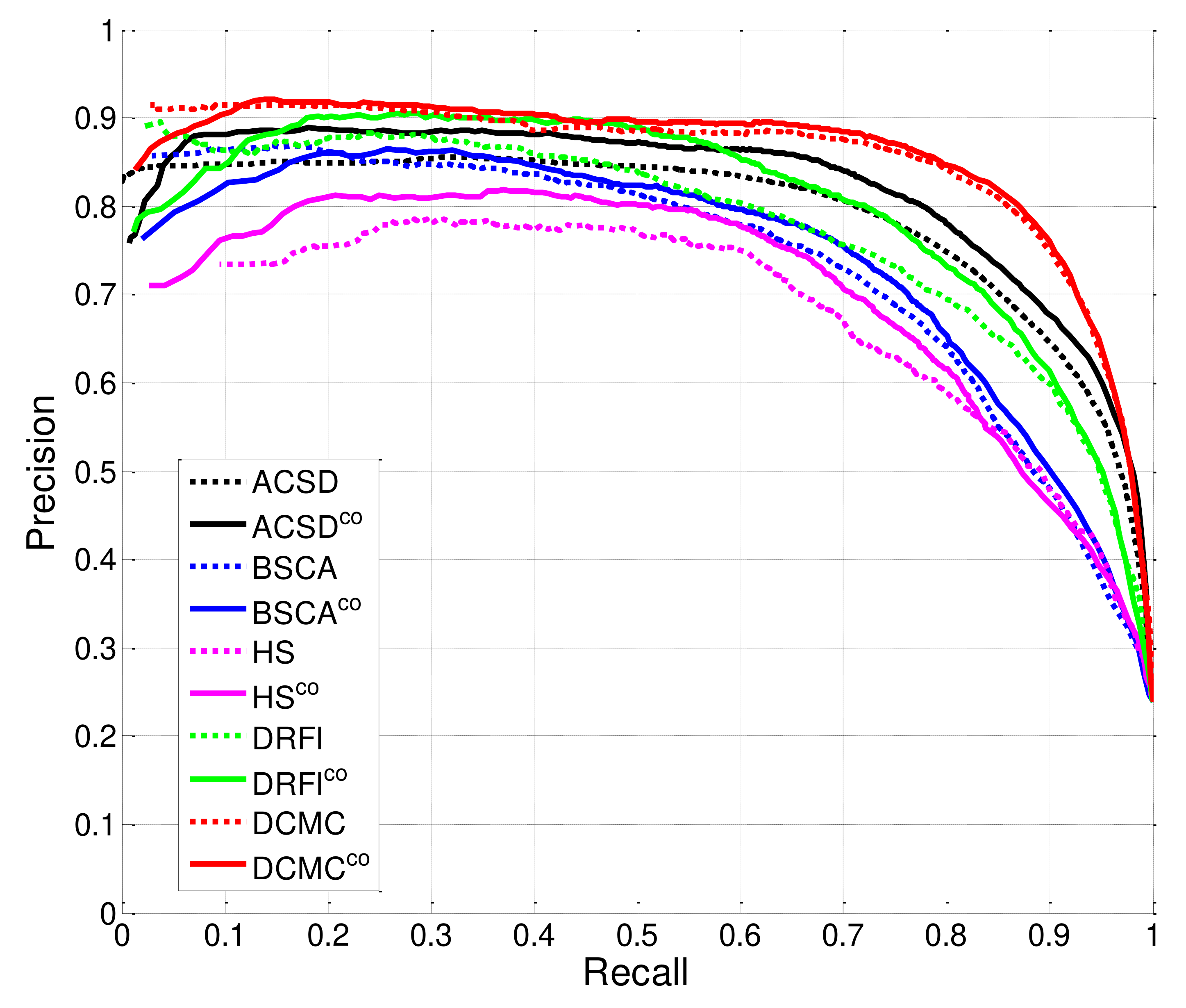}
\caption{The PR curves of co-saliency results on the RGBD Cosal150 dataset with different intra saliency maps. The subscript of ``co'' means the co-saliency result produced by our co-saliency model with CLP optimization and different intra saliency maps. }
\label{fig7}
\end{figure}
\begin{table}[!b]
\renewcommand\arraystretch{1.8}
\centering
\caption{Quantitative Evaluations of Co-saliency Results with Different Intra Saliency Models on the RGBD Cosal150 Dataset}
\begin{tabular}{c|c|c|c|c}
\toprule[1.5pt]
 \multicolumn{2}{c|}{} & F-measure & MAE & Area under PR curves \\[0.5ex]
\hline
\multirow{2}{*}{ACSD}&intra saliency & $0.7788$ & $0.1806$ & $0.7915$ \\
  & co-saliency & $\bm{0.8039}$ & $\bm{0.1529}$ & $\bm{0.8154}$ \\
\hline
\multirow{2}{*}{BSCA}&intra saliency & $0.7318$ & $0.1925$ & $0.7165$ \\
  & co-saliency & $\bm{0.7470}$ & $\bm{0.1731}$ & $\bm{0.7279}$ \\
\hline
\multirow{2}{*}{HS}&intra saliency & $0.7101$ & $0.2375$ & $0.6131$ \\
  & co-saliency & $\bm{0.7283}$ & $\bm{0.2063}$ & $\bm{0.6878}$ \\
\hline
\multirow{2}{*}{DRFI}&intra saliency & $0.7484$ & $0.1949$ & $0.7640$ \\
  & co-saliency & $\bm{0.7814}$ & $\bm{0.1642}$ & $\bm{0.7935}$ \\
\hline
\multirow{2}{*}{DCMC}&intra saliency & $0.8348$ & $0.1498$ & $0.8295$ \\
  & co-saliency & $\bm{0.8403}$ & $\bm{0.1370}$ & $\bm{0.8472}$ \\
\bottomrule[1.5pt]
\end{tabular}
\label{tab4}
\end{table}
In this paper, we focus on designing an opened framework to make the existing saliency detection methods work well in co-saliency scenarios. Therefore, the experiments are conducted to evaluate the performance of our model with different intra saliency maps on the RGBD Cosal150 dataset. In the experiment, five different single saliency maps produced by ACSD \cite{R33}, HS \cite{R40}, BSCA \cite{R41}, DRFI \cite{R53}, and DCMC \cite{R36} are used as the intra saliency maps. The PR curves are illustrated in Fig. \ref{fig7}, and some quantitative indictors including the F-measure, MAE score, and the area under PR curves are reported in TABLE \ref{tab4}. In Fig. \ref{fig7}, the PR curves of the results using our co-saliency framework are higher than those from the original saliency maps. The consistent conclusion can be drawn from the quantitative comparisons in TABLE \ref{tab4}. The F-measure is improved through our proposed co-saliency model, and the MAE scores and area under the PR curves also achieve better performances compared with the previous saliency results. Taking the F-measure as an example, the proposed co-saliency model achieves a maximum percentage gain of 4.41\% compared to the corresponding intra saliency result, and the average percentage gain also reaches 2.59\%. Similarly, the maximum percentage gains of the MAE scores and area under PR curves achieve 15.75\% and 12.18\%, respectively. Moreover, in general, the better the single saliency map (intra saliency map) achieves, the better performance of the co-saliency map is. In brief, the results demonstrate that the proposed model can effectively improve the performance of the existing single saliency models, and make them work well for co-saliency detection. \par

\subsection{Discussion}
\begin{figure}[htbp]
\centering
\includegraphics[width=1\linewidth]{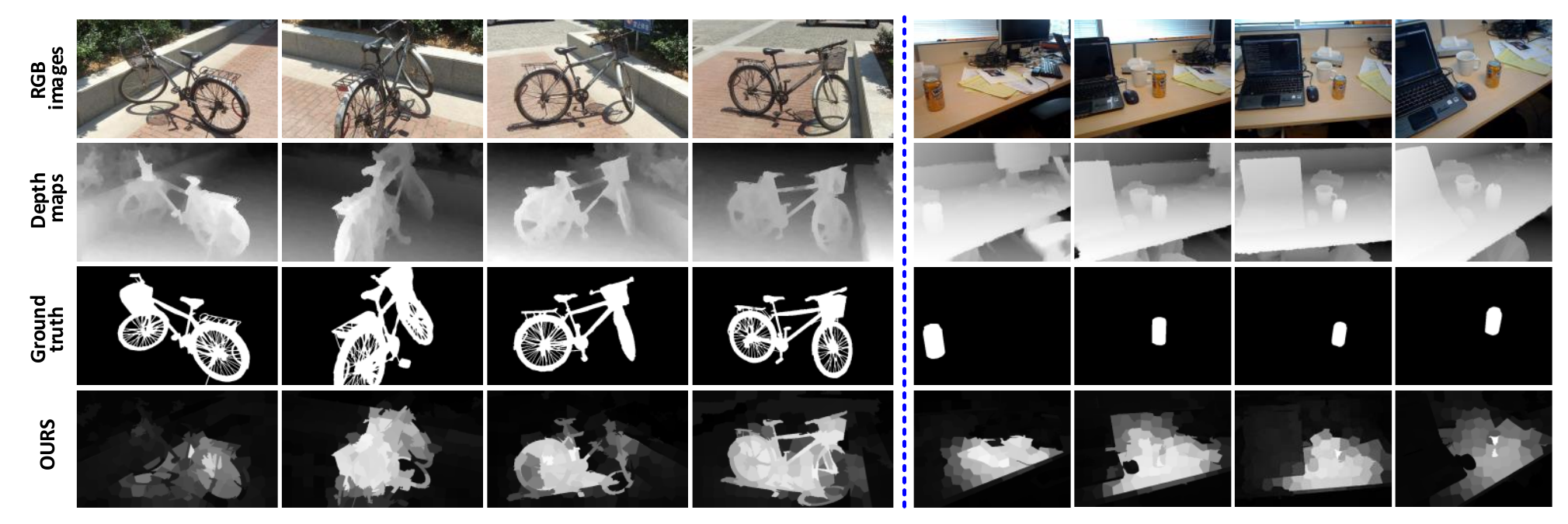}
\caption{Some challenging examples for our RGBD co-saliency detection model. }
\label{fig8}
\end{figure}

We discuss some challenging cases of the proposed RGBD co-saliency model in this subsection. The details are shown in Fig. \ref{fig8}. For the bike group, the salient foreground is very trivial and includes lots of stuff regions, such as the spokes and back seat. These regions are difficult to detect completely and accurately through our co-saliency model. For the soda can group, the scene is relatively complex and cluttered, and the soda can is too small to be detected as the salient object compared with the computer in each image. Thus, the small scale object is not detected successfully through our model, especially in the complex and cluttered scenes. \par
\section{Conclusion}
In this paper, a co-saliency detection model for RGBD images is presented, which focuses on mining the constraint relationship among multiple images and introducing the depth cue into the co-saliency model. The similarity constraint, saliency consistency, and cluster-based constraint are introduced in feature similarity matching to obtain more stable and accurate corresponding relationship at superpixel level. The image-level similarity descriptor is designed as the weighted coefficient for inter saliency calculation. In addition, a CLP optimization strategy is proposed to optimize the intra and inter saliency maps in a cross way. The comprehensive comparisons on two RGBD co-saliency detection datasets have demonstrated that the proposed method outperforms other state-of-the-art saliency and co-saliency models. It also verifies the effectiveness of improving the existing saliency models in co-saliency application.\par

\ifCLASSOPTIONcaptionsoff
  \newpage
\fi

\end{document}